\documentclass{article}
\usepackage{nips14submit_e,times}
\usepackage{mathtools}
\usepackage{hyperref}
\usepackage{url}
\usepackage{algorithm}
\usepackage{algpseudocode}
\nipsfinalcopy

\usepackage[margin=1in]{geometry}

\usepackage{amsmath}
\usepackage{amssymb}

\usepackage{mario-def}

\usepackage{color}

\title{On the String Kernel Pre-Image Problem with Applications in Drug Discovery}

\begin{document}

\maketitle

{\vspace{-20mm} \large
\begin{center}
S{\'e}bastien Gigu{\`e}re, Am{\'e}lie Rolland\footnote{Corresponding author: \href{mailto:amelie.rolland.1@ulaval.ca}{amelie.rolland.1@ulaval.ca}\\{\indent Peer-reviewed and accepted for presentation at Machine Learning in Computational Biology 2014, Montr\'{e}al, Qu\'{e}bec, Canada.}}, Fran{\c c}ois Laviolette, Mario Marchand
\end{center}}

\begin{center}
\begin{tabular}{cc}
\multicolumn{2}{c}{\large Universit{\'e} Laval}\\
 & \\
D{\'e}partement d'informatique et de g{\'e}nie logiciel 
\end{tabular}
\end{center}

\vspace{7mm}
\begin{abstract}
The pre-image problem has to be solved during inference by most structured output predictors.
For string kernels, this problem corresponds to finding the string associated to a given input.
An algorithm capable of solving or finding good approximations to this problem would have many applications in computational biology and other fields.
This work uses a recent result on combinatorial optimization of linear predictors based on string kernels to develop, for the pre-image, a low complexity upper bound valid for many string kernels.
This upper bound is used with success in a branch and bound searching algorithm. 
Applications and results in the discovery of druggable peptides are presented and discussed.
\end{abstract}

\section{Introduction}
This work focuses on the challenges of using regression learning algorithms in the design of highly active peptides.
Let $\Acal$ be the set of all amino acids and $\Acal^*$ be the set of all possible peptides.
Throughout this paper we will assume that we have a dataset $\Scal = \{(\xb_1, e_1),\ldots, (\xb_m, e_m) \} \in \Acal^* \times \mathbb{R}$ where $\xb_i$ is the amino acid sequence of the $i$-th peptide in $\Scal$ and $e_i$ is its bioactivity (which could be its binding affinity to some target protein, its antimicrobial activity, or some other desirable activity for peptides).  
The regression approach consists of learning a predictor $h$ from the training dataset $\Scal$.
Let $h(\xb)$ be the estimated bioactivity of $\xb$ according to the predictor $h$.
There are many ways to represent such a predictor but this work focuses on string kernel based predictors.
Many learning algorithm, such as the Support Vector Regression, Ridge Regression, and Gaussian Processes, produce predictors $h$ whose output $h(\xb)$, on input $\xb$, is given by
\begin{equation}\label{eq:unnormalized_predictor}
h(\xb) = \sum_{i=1}^m \alpha_i K(\xb_i,\xb) \, ,
\end{equation}
where $\alpha_i$ is the weight on the $i$-th training example, $K(\xb_i,\xb) = \langle \phib(\xb), \phib(\xb') \rangle$ is a string kernel, $\langle \cdot, \cdot \rangle$ denotes the inner product between feature vectors, and $\phib$ is the feature map associated to $K$.
The weight vector $\alb=(\alpha_1,\alpha_2,.., \alpha_m)$ is obtained by minimizing the learning algorithm objective function. 

For the discovery of peptide inhibitors and peptides that could act as drug precursors, we are interested in finding the peptide $\xb^h$ maximizing the predicted bioactivity:
\begin{equation}\label{arg_max_unnormalized}
\xb^h = \arg \max_{\xb \in \Acal^*} h(\xb)\, .
\end{equation}
The complexity of this combinatorial problem will obviously depend on the string kernel used to learn $h$.
For few kernels like the Hamming kernel, solving Equation~\eqref{arg_max_unnormalized} is trivial~\cite{Giguere2013Risk}. However, this kernel is not well suited for peptides~\cite{Giguere2013Learning}.
On another hand, the Generic String (GS) kernel~\cite{Giguere2013Learning} is well suited for peptides. It is defined as follows:
\begin{equation}\label{eq:GS_kernel}
GS(\xb,\xb', k, \sigma_p, \sigma_c) \eqdef \sum_{l=1}^{k}\sum_{i=0}^{|x|-l} \sum_{j=0}^{|x'|-l}\,  \exp{\left(\mbox{\large $\frac{-(i - j)^2}{2\sigma_p^2}$}\right)} \exp{\left(\mbox{\large $\frac{-\parallel \psib^l(x_{i+1},..,x_{i+l}) \,-\, \psib^l(x'_{j+1},..,x'_{j+l}) \parallel^2}{2\sigma_c^2}$}\right)}\, ,
\end{equation}
where $k$ controls the length of compared $k$-mers, \,$\psib^k: \Acal^k\rightarrow\mathbb{R}^{dk}$ encodes the physico-chemical properties of $k$-mers by mapping each of the $k$ amino acids to a real valued vector containing $d$ properties, $\sigma_c$ controls the penalty incurred when the physico-chemical properties of two $k$-mers differ, and $\sigma_p$ controls the penalty incurred when two $k$-mers are not sharing the same position in their respective peptides. 
Depending on the chosen hyper-parameters, this kernel can be specialized to eight known kernels \cite{Giguere2013Learning}, namely the Hamming kernel, the Blended Spectrum \cite{Shawe2004Kernel}, the Radial Basis Function (RBF), the Oligo \cite{Meinicke2004Oligo}, and the Weighted degree \cite{Ratsch2004Accurate}.
Since the proposed approach uses the GS kernel it is also valid for all of these kernels.

It was recently shown \cite{giguere2014Machine} that when $K$ is the Generic String (GS) kernel, and when we restrict peptides to be of length $l$, the peptide $\xb^h \in \Acal^l$ can be found in polynomial time for any predictor $h$ in the form of Equation~\eqref{eq:unnormalized_predictor}.
Their approach maps the combinatorial problem to a directed acyclic graph (DAG) that is basically a \emph{de} Bruijn graph with weights on the arcs.
Then, they show that finding the longest (weighted) path in this graph is equivalent to finding $\xb^h \in \Acal^l$.
Finally, the longest path can be found in polynomial time by dynamic programming since the graph is acyclic.

However, with the GS kernel, given two peptides $\xb$ and $\xb'$ of the same length, the Euclidean norms of their feature vectors 
$\phib(\xb)$ and $\phib(\xb')$ can differ substantially, i.e. we can have $\sqrt{K(\xb, \xb)} \gg  \sqrt{K(\xb', \xb')}$. Let us first show, with a simple example, why this can be problematic. We will then show how to avoid this problem through kernel normalization.
Such normalization does not have any computational impact in the learning phase of the regression algorithm.
It does however impact substantially the prediction phase, leading to a harder combinatorial problem.

Hence, let us consider the strings ``AAAAA'' and ``ABCDE'' and the Spectrum kernel~\cite{Leslie2002Spectrum} (also known as the $n$-gram kernel) with $k$-mers of size two.
The string ``AAAAA'' has a norm of $\sqrt{16}=4$, while ``ABCDE'' has a norm of $\sqrt{1+1+1+1} = 2$.
Hence the Spectrum kernel is sensitive to $k$-mers repetitions in the string.  Note that this problem is shared by many other string kernels.
In the case of the GS kernel, it is possible to avoid this problem by fixing the hyper-parameter $\sigma_p$ to $0$, which forces a constant norm, i.e. $K(\xb,\xb) = c$ for all $\xb \in \Acal^l$.
However, $K(\xb,\xb)$ will vary whenever $\sigma_p > 0$, as it is the case for the Blended Spectrum and the Oligo kernels, for example.
A consequence of a non-constant norm is that $h(\xb)$ will heavily depend on the norm of $\phib(\xb)$. Hence,   $\xb^h$ will be more likely a peptide having a feature vector with a large norm, i.e., a peptide having many repetitions. This is generally an undesired bias.
It is easy to overcome this problem by normalizing kernel values, in that case the output function becomes
\begin{equation}\label{eq:normalized_predictor}
h_\star(\xb) = \sum_{i=1}^m \alpha_i \frac{K(\xb_i,\xb)}{\sqrt{K(\xb_i,\xb_i) K(\xb,\xb)}}
= \frac{1}{\sqrt{K(\xb,\xb)}} \sum_{i=1}^m \beta_i K(\xb_i,\xb)
\end{equation}
where $\beta_i = \frac{\alpha_i}{\sqrt{K(\xb_i,\xb_i)}}$\, .
In that case we are now interested in finding
\begin{equation}\label{arg_max_normalized}
\xb^{h_\star} = \arg \max_{\xb \in \Acal^*} \frac{1}{\sqrt{K(\xb,\xb)}} \sum_{i=1}^m \beta_i K(\xb_i,\xb) \, .
\end{equation}
This optimization problem is also a pre-image problem, but written in a slightly different form.
We conjecture that solving Equation~\eqref{arg_max_normalized} is $\Ncal\Pcal$-Hard when $K$ is the GS kernel.
Given the similarity between the problems of Equation~\eqref{arg_max_unnormalized} and Equation~\eqref{arg_max_normalized}, the difference in their computational complexity is unexpected.

In the next section, we will present a low complexity upper bound on Equation~\eqref{eq:normalized_predictor} that makes it a good candidate for a branch and bound search to solve Equation~\eqref{arg_max_normalized}.

\section{Method}
Since the number of peptides grows exponentially with its length, it becomes impossible to evaluate all solutions for large peptides, we propose a branch and bound scheme to guide this search.
A branch and bound algorithm starts by dividing the search space into disjoint subspaces.
For example, one subspace could be all peptides ending with the string ``DE''.
For a maximization problem, an upper bound on the best achievable solution is computed for each of these subspaces.
Then, the subspace with the highest upper bound is further divided into smaller subspaces.
Finally, the search stops when a subspace can no longer be divided (a leaf is reached in the search tree), or when the upper bound value is lower than the value of an already achieved solution (\emph{i.e.}, an already reached leaf in the search tree).
A branch and bound approach can thus avoid exploring a large part of the search space.

Algorithm~\ref{algo:bandb} gives the specifics of the branch and bound algorithm applied to our case.
The search algorithm alternates between a greedy phase and a branch and bound phase.
The greedy phase is important to ensure that leaves of the search tree are quickly visited.
This allows good but sub-optimal solutions to be returned by the algorithm if the allowed computational time expires.
Whenever a node is visited, the bound is computed for all its children and they are added to the priority queue accordingly.
This greedy process is repeated until a leaf is reached.
Then, the node with the largest bound is visited and the greedy process starts again.
At all time, the best solution found so far is kept and the search stops when the bound of the node on top of the priority queue is smaller than the value of the best solution.

\begin{algorithm}
\caption{Branch and bound search for maximal string of length $l$ }
\label{algo:bandb}
\begin{algorithmic}
    \State $\Qcal$ : empty priority queue ordering bounds in descending order
    \\
    \State $best\_node \gets \textit{Node}(\textit{empty\_string}, 0)$
    \ForAll{$s\in\Acal^k$}
    	\Comment Add all $k$-mer in $\Qcal$
    	\State $\Qcal.push(\textit{Node}(s, F(s, l)))$
    \EndFor
    \While{$current\_node \gets \Qcal.pop() \And current\_node.bound > best\_node.bound$}
    	\Comment Get maximal node in $\Qcal$
		\While{$|current\_node.string| < l \And current\_node.bound > best\_node.bound$}
			 \State $best\_child \gets \textit{Node}(\textit{empty-string}, 0)$
			 \ForAll{$a \in \Acal$}
			 	\Comment Evaluate all children of node
			 	\State $s' \gets \textit{Concatenate}(a, current\_node.string)$
			 	\If{$F(s',l) > best\_node.bound$}
			 		\If{$F(s',l) > best\_child.bound$}
						
						\State $best\_child \gets \textit{Node}(s', F(s',l))$
						\Comment Update $best\_child$
			 		\EndIf
					\State $\Qcal.push(\textit{Node}(s', F(s',l)))$	
					\Comment Add child to $\Qcal$
			 	\EndIf
    	\EndFor
    	\State $current\_node \gets best\_child$
    	\State $\Qcal.remove(best\_child)$
    	\Comment Remove $best\_child$ from $\Qcal$ if it was added
		\EndWhile
		\If{$|current\_node.string|=l \And current\_node.bound > best\_node.bound$}
			\State $best\_node \gets current\_node$
			\Comment Update $best\_node$
		\EndIf
    \EndWhile
    \\
\\
\Return $best\_node.string$, $best\_node.bound$
\Comment Return string and maximal bioactivity

\end{algorithmic}
\end{algorithm}

Let $\Acal^{l-p} \times \{x'_1, \ldots, x'_p \}$ be the set of all possible strings of length $l$ that end with $x'_1, \ldots, x'_p$, in other words, $\Acal^{l-p} \times \{x'_1, \ldots, x'_p \}$ is a set of strings having their last $p$ characters fixed.
Our goal is to have a function $F$ that upper bounds $h_\star$ for every $\Acal^{l-p} \times \{x'_1, \ldots, x'_p \}$. In other words:
\begin{equation}
F(\xb', l) \geq \max_{\xb \in \Acal^{l-p} \times \{x'_1, \ldots, x'_p \}} \frac{1}{\sqrt{K(\xb,\xb)}} \sum_{i=1}^m \beta_i K(\xb_i,\xb) \,.
\end{equation}
To do so, let $F(\xb', l) \eqdef \frac{1}{\sqrt{f(\xb', l)}} g(\xb', l)$, where
\begin{align}
f(\xb', l) \leq \min_{\xb \in \Acal^{l-p} \times \{x'_1, \ldots, x'_p \}} K(\xb,\xb) &&\hbox{and}&&
g(\xb', l) \geq \max_{\xb \in \Acal^{l-p} \times \{x'_1, \ldots, x'_p \}} \sum_{i=1}^m \beta_i K(\xb_i,\xb)
\end{align}
are, respectively, a lower and  an upper bound.

When $K$ is the GS kernel with hyper-parameters $k$ , $\sigma_p$ and  $\sigma_c$, the  lower bound $f$ can be obtained as follows: 
\begin{equation}\label{eq:f_min}
f_{\mbox{$_{k, \sigma_p, \sigma_c}$}}
(\xb', l) \eqdef GS(\xb',\xb', k, \sigma_p, \sigma_c) + 2XX'(\xb', l, k, \sigma_p, \sigma_c) + XX(\xb', l, k, \sigma_p, \sigma_c)\,,
\end{equation}
where
\begin{equation}\label{eq:f_xxprime}
XX'(\xb', l, k, \sigma_p, \sigma_c)\eqdef \sum_{p=1}^{k} \sum_{i=1}^{l-|\xb'|} \min_{\xb \in \Acal^p} \sum_{j=1}^{|\xb'|}\, \exp{\left(\mbox{\large $\frac{-(i - j)^2}{2\sigma_p^2}$}\right)} \exp{\left(\mbox{\large $\frac{-\parallel \psib^l(\xb_1,..,\xb_p) \,-\, \psib^l(x'_{j},..,x'_{j+p}) \parallel^2}{2\sigma_c^2}$}\right)}\,,
\end{equation}
\begin{equation}\label{eq:f_xx}
XX(\xb', l, k, \sigma_p, \sigma_c) \eqdef \sum_{p=1}^{k}\sum_{i=1}^{l-|\xb'|} \sum_{j=1}^{l-|\xb'|}\,  \exp{\left(\mbox{\large $\frac{-(i -(l-|\xb'|+j))^2}{2\sigma_p^2}$}\right)} \exp{\left(\mbox{\large $\frac{- (D(i, j)^2 + ... + D(i+p, j+p)^2}{2\sigma_c^2}$}\right)}\,
\end{equation}
and 
\[
D(i, j) = \begin{dcases*}
        0 & if $i=j$,\\
        \max_{a,a' \in \Acal} \psib(a, a') & otherwise.  \\
        \end{dcases*}
\]
Note that the lower bound $f$ is not attained since it under-estimates the value of the string $\xb \in \Acal^{l-p} \times \{x'_1, \ldots, x'_p \}$ minimizing $K(\xb, \xb)$.

For $g(\xb', l)$, note that $\sum_{i=1}^m \beta_i K(\xb_i,\xb)$ is what \cite{giguere2014Machine} proposed to maximize.
Their approach uses a dynamic programming table to compute the longest path in a graph.
It is relatively easy to modify their algorithm to return the table instead of the longest path.
In that way, given a string with suffix $\xb'$, it is possible to determine in constant time, by accessing the dynamic programming table, the prefix from $\Acal^{l-p}$ maximizing $\sum_{i=1}^m \beta_i K(\xb_i,\xb)$.
The computation of $g$ is thus very efficient, the algorithm of \cite{giguere2014Machine} only needs to be done once before the branch and bound search, then $g(\xb', l)$ can be computed in constant time for any $\xb'$. Finally, $g$ is the least upper bound (or suprema) since there is always a string $\xb\in \Acal^{l-p} \times \{x'_1, \ldots, x'_p \}$ with $g(\xb,l) = \sum_{i=1}^m \beta_i K(\xb_i,\xb)$.
In other words, there are no tighter bound.

\section{Results and Discussion}
We followed the protocol suggested by \cite{giguere2014Machine} and, as a proof of concept, we used the same dataset they used: $101$ cationic antimicrobial pentadecapeptides (CAMPs) from the synthetic antibiotic peptides database \cite{Wade2002Synthetic}.
Peptide antibacterial activities are expressed as the logarithm of bactericidal potency.
As in \cite{giguere2014Machine}, we used kernel ridge regression as the learning algorithm.
Except when stated otherwise, all hyper-parameters for the GS kernel ($k,\sigma_c, \sigma_p$) and the kernel ridge regression ($\lambda$) were chosen by standard cross-validation.
We learned two predictors of antimicrobial potency, one uses unnormalized kernel values (thus, the same predictor used in \cite{giguere2014Machine}), the other was trained using normalized kernel values.
We refer to these predictors respectively as $h$ and $h_\star$.

The method presented in \cite{giguere2014Machine} was used to identify the peptide $\xb^h \in \Acal^{15}$ of maximal predicted bioactivity according to $h$ and the branch and bound was used to identify $\xb^{h_\star} \in \Acal^{15}$, the peptide maximizing $h_\star$.
Both approaches found the same peptide, which is ``WWKWWKRLRRLFLLV''.

Note that both methods are able to output more than one sequence.
The method of \cite{giguere2014Machine} uses a $k$-longest path algorithm to obtain the $k$ peptides of maximal bioactivity.
It is also possible for the branch and bound by stopping the search only when the bound on top of the priority queue is lower than the $k$-th peptide found.
These peptides can be used, for example, for motif generation, for multiple peptide synthesis or in combinatorial chemistry.
To highlight the differences between the methods, they were used to list the top $1000$ peptides and the lists were compared: $70.5 \%$ of the $1000$ peptides were found by both methods.
Then, the Pearson correlation coefficient (PCC) was computed between the rank of the $679$ peptides present in both lists: a PCC of $0.56$ was obtained.
Hence, the ranking of peptides found by $h$ and by $h_\star$ differ significantly.

The overlap in the lists is attributed to the value of $\sigma_p$ that was chosen during cross-validation for $h$.
As explained earlier, when $\sigma_p$ is $0$, the unnormalized predictor $h$ will not suffer from differences in the norm of $\phib(\xb)$.
In the case of antimicrobial peptides, $\sigma_p=0.8$ was found to be the optimal value for $h$.
This suggests that the method of~\cite{giguere2014Machine} offers some resistance to variation in norm when $\sigma_p$ is small. Indeed, for small $\sigma_p$, all examples have about the same norm.

To further highlight the difference between the methods, we intentionally fixed $\sigma_p$ to infinity and cross-validated all other parameters for $h$ and $h_\star$ and compared the best peptides found.
Situations were $\sigma_p$ would have to be set to infinity are not at all unlikely. For example, cyclic peptides have no N-terminus and no C-terminus. For that reason, there is no origin from which we can express the positions of $k$-mers in these peptides.
The method of \cite{giguere2014Machine} found the peptide ``FKKIFKKIFKKIFKF'' using the predictor~$h$ and the branch and bound approach found the peptide ``WKKIFKKIWKFRVFK'' using the predictor $h_\star$.
The peptide identified by $h$ shares almost no similarity with peptides of the training set and is basically composed of repetition of the $k$-mer ``FKK''.
In contrast, the peptide identified by $h_\star$ shares many substructures with the most bioactive peptides of the training set. This tends to point out that in situation where example norms vary a lot, $h$ clearly favors examples having a large norm.
However, this bias is unjustified and is not related to a biological reality. 

\section{Conclusion}
We proposed a bound for maximizing the inference function of kernel methods that uses normalized string kernels.
Moreover, the bound is also valid for solving the pre-image of a variety of string kernels.
Empirical results show that the method proposed by \cite{giguere2014Machine} can suffer from a dominance of the norm for certain strings, a problem which is present with many string kernels.
In these situations, the proposed method was shown to overcome this problem.
Tighter bounds should take advantage of the proposed framework and allow the discovery of novel peptide inhibitors.
Finally, applications for drugs based on cyclic peptides and other structured output applications are expected.

\subsubsection*{Acknowledgments}

The authors would like to thank Prudencio Tossu for his help in the experimentations. Computations were performed on the Colosse supercomputer at Universit\'{e} Laval (resource allocation project: avt-710-aa), under the auspices of Calcul Qu\'{e}bec and Compute Canada. AR is recipient of a Master's Scholarship from the Fonds de recherche du Qu\'{e}bec - Nature et technologies (FRQNT). This work was supported by the FRQNT (FL \&  MM; 2013-PR-166708).

\bibliography{these_seb}

\begin{thebibliography}{1}

\bibitem{Giguere2013Risk}
S{\'e}bastien Gigu{\`e}re, Fran\c{c}ois Laviolette, Mario Marchand, and
  Khadidja Sylla.
\newblock Risk bounds and learning algorithms for the regression approach to
  structured output prediction.
\newblock In {\em International Conference on Machine Learning (ICML)}, 2013.

\bibitem{Giguere2013Learning}
S\'ebastien Gigu\`ere, Mario Marchand, Fran\c{c}ois Laviolette, Alexandre
  Drouin, and Jacques Corbeil.
\newblock Learning a peptide-protein binding affinity predictor with kernel
  ridge regression.
\newblock {\em BMC Bioinformatics}, 14, 2013.

\bibitem{Shawe2004Kernel}
John Shawe-Taylor and Nello Cristianini.
\newblock {\em Kernel methods for pattern analysis}.
\newblock Cambridge university press, 2004.

\bibitem{Meinicke2004Oligo}
P.~Meinicke, M.~Tech, B.~Morgenstern, and R.~Merkl.
\newblock Oligo kernels for datamining on biological sequences: A case study on
  prokaryotic translation initiation sites.
\newblock {\em BMC Bioinformatics}, 5, 2004.

\bibitem{Ratsch2004Accurate}
Gunnar R{\"{a}}tsch and S{\"{o}}ren Sonnenburg.
\newblock {Accurate Splice Site Detection for Caenorhabditis elegans}.
\newblock In {B} and J.~P. Vert, editors, {\em Kernel Methods in Computational
  Biology}, pages 277--298. MIT Press, 2004.

\bibitem{giguere2014Machine}
S{\'e}bastien Gigu{\`e}re, Fran{\c{c}}ois Laviolette, Mario Marchand, Denise
  Tremblay, Sylvain Moineau, {\'E}ric Biron, and Jacques Corbeil.
\newblock Machine learning assisted design of highly active peptides for drug
  discovery.
\newblock {\em Under review, Submitted to MLCB}, 2014.

\bibitem{Leslie2002Spectrum}
Christina~S Leslie, Eleazar Eskin, and William~Stafford Noble.
\newblock The spectrum kernel: A string kernel for svm protein classification.
\newblock In {\em Pacific symposium on biocomputing}, volume~7, pages 566--575.
  World Scientific, 2002.

\bibitem{Wade2002Synthetic}
David Wade and Jukka Englund.
\newblock Synthetic antibiotic peptides database.
\newblock {\em Protein and peptide letters}, 9(1):53--57, 2002.

\end{thebibliography}
\bibliographystyle{unsrt}
\end{document}